\documentclass[letterpaper]{article} 
\usepackage{aaai23}  
\usepackage{times}  
\usepackage{helvet}  
\usepackage{courier}  
\usepackage[hyphens]{url}  
\usepackage{graphicx} 
\usepackage{natbib}  
\usepackage{caption} 
\usepackage{algorithm}
\usepackage{algorithmic}
\usepackage{amsmath,amssymb} 
\usepackage{color}
\usepackage{booktabs}
\usepackage{multirow}
\usepackage{subfig}
\usepackage[colorlinks,linkcolor=red,citecolor=black]{hyperref}

\usepackage{newfloat}
\usepackage{listings}
\DeclareCaptionStyle{ruled}{labelfont=normalfont,labelsep=colon,strut=off} 
\floatstyle{ruled}
\newfloat{listing}{tb}{lst}{}
\floatname{listing}{Listing}
\title{SepViT: Separable Vision Transformer}
\author {
    Wei Li \textsuperscript{\rm 1,2,$\dagger$}\thanks{Corresponding Authors. $\dagger$Intern at ByteDance. \\
    \indent \ \ \ Email: liwei9719@126.com, wangxing.613@bytedance.com}, 
    Xing Wang\textsuperscript{\rm 2,$\ast$}, 
    Xin Xia\textsuperscript{\rm 2}, 
    Jie Wu\textsuperscript{\rm 2}, \\
    Jiashi Li\textsuperscript{\rm 2}, 
    Xuefeng Xiao\textsuperscript{\rm 2}, 
    Min Zheng\textsuperscript{\rm 2}, 
    Shiping Wen\textsuperscript{\rm 3}
}
\affiliations {
    \textsuperscript{\rm 1} University of Electronic Science and Technology of China\\
    \textsuperscript{\rm 2} ByteDance Inc, 
    \textsuperscript{\rm 3} University of Technology Sydney
}

\usepackage{bibentry}

\begin{document}

\maketitle

\begin{abstract}
Vision Transformers have witnessed prevailing success in a series of vision tasks. However, these Transformers often rely on extensive computational costs to achieve high performance, which is burdensome to deploy on resource-constrained devices. To alleviate this issue, we draw lessons from depthwise separable convolution and imitate its ideology to design an efficient Transformer backbone, i.e., \emph{Separable Vision Transformer}, abbreviated as \emph{SepViT}. SepViT helps to carry out the local-global information interaction within and among the windows in sequential order via a depthwise separable self-attention. The novel window token embedding and grouped self-attention are employed to compute the attention relationship among windows with negligible cost and establish long-range visual interactions across multiple windows, respectively. Extensive experiments on general-purpose vision benchmarks demonstrate that SepViT can achieve a state-of-the-art trade-off between performance and latency. Among them, SepViT achieves 84.2\% top-1 accuracy on ImageNet-1K classification while decreasing the latency by 40\%, compared to the ones with similar accuracy (e.g., CSWin). Furthermore, SepViT achieves 51.0\% mIoU on ADE20K semantic segmentation task, 47.9 AP on the RetinaNet-based COCO detection task, 49.4 box AP and 44.6 mask AP on Mask R-CNN-based COCO object detection and instance segmentation tasks. The code is available at: \url{https://github.com/liwei109/SepViT}.
\end{abstract}

\section{Introduction}
Recently, many computer vision researchers make efforts to design CV-oriented vision Transformer to surpass the performance of the convolutional neural networks (CNNs).
Due to a high capability in modeling the long-range dependencies, vision Transformer achieves prominent results in diversified vision tasks, such as image classification \cite{ViT,Deit,TNT,Swin,Twins,CSWin}, semantic segmentation \cite{Segformer,zheng2021rethinking,wang2021max}, object detection \cite{DETR,zhu2020deformable,dai2021up} and etc. However, the powerful performance usually comes at a cost of heavy computational complexity.

\begin{figure*}[th]
  \centering
  \subfloat{\includegraphics[width=0.35\textwidth]{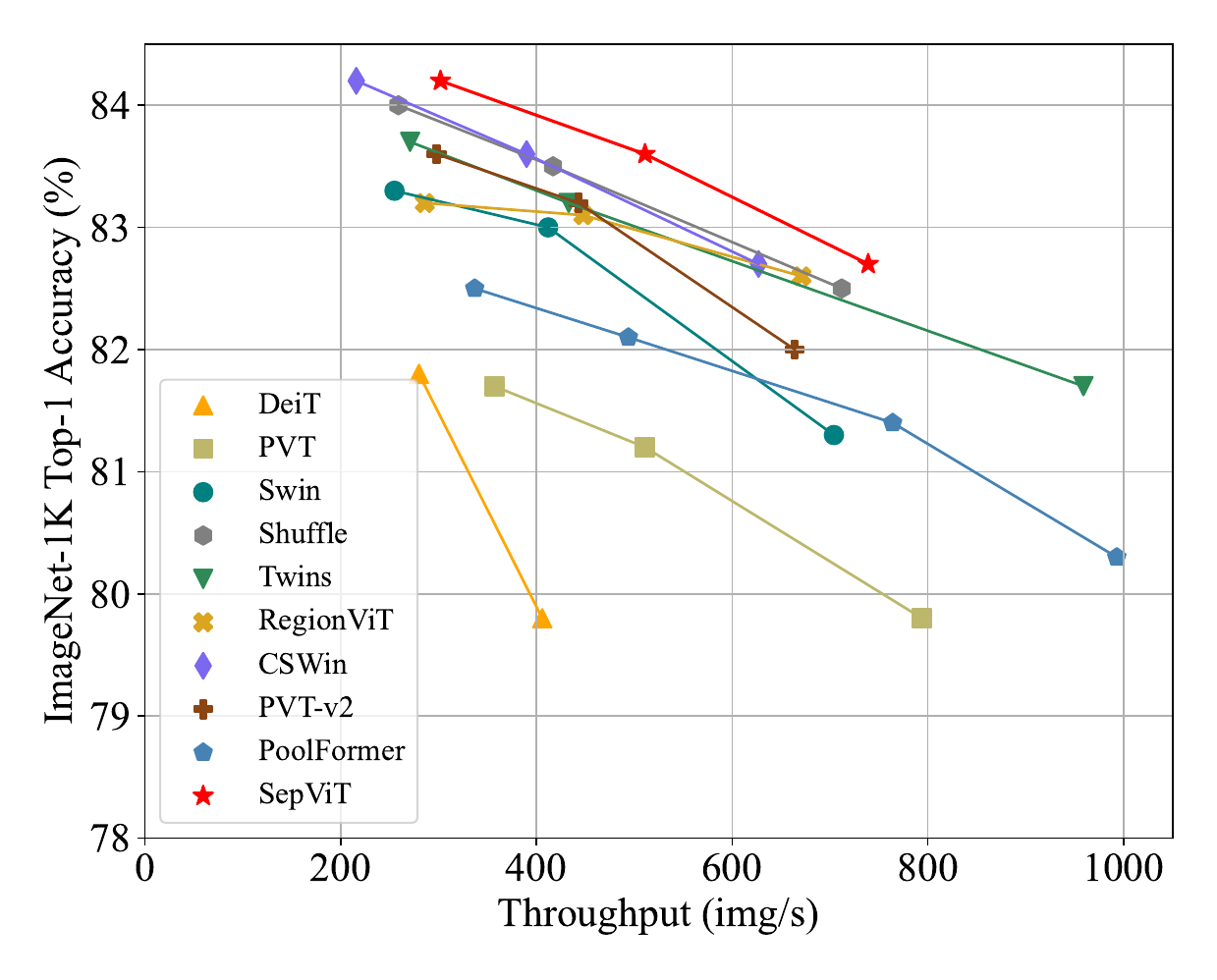}\label{latency}}
  \hspace{5mm}
  \subfloat{\includegraphics[width=0.35\textwidth]{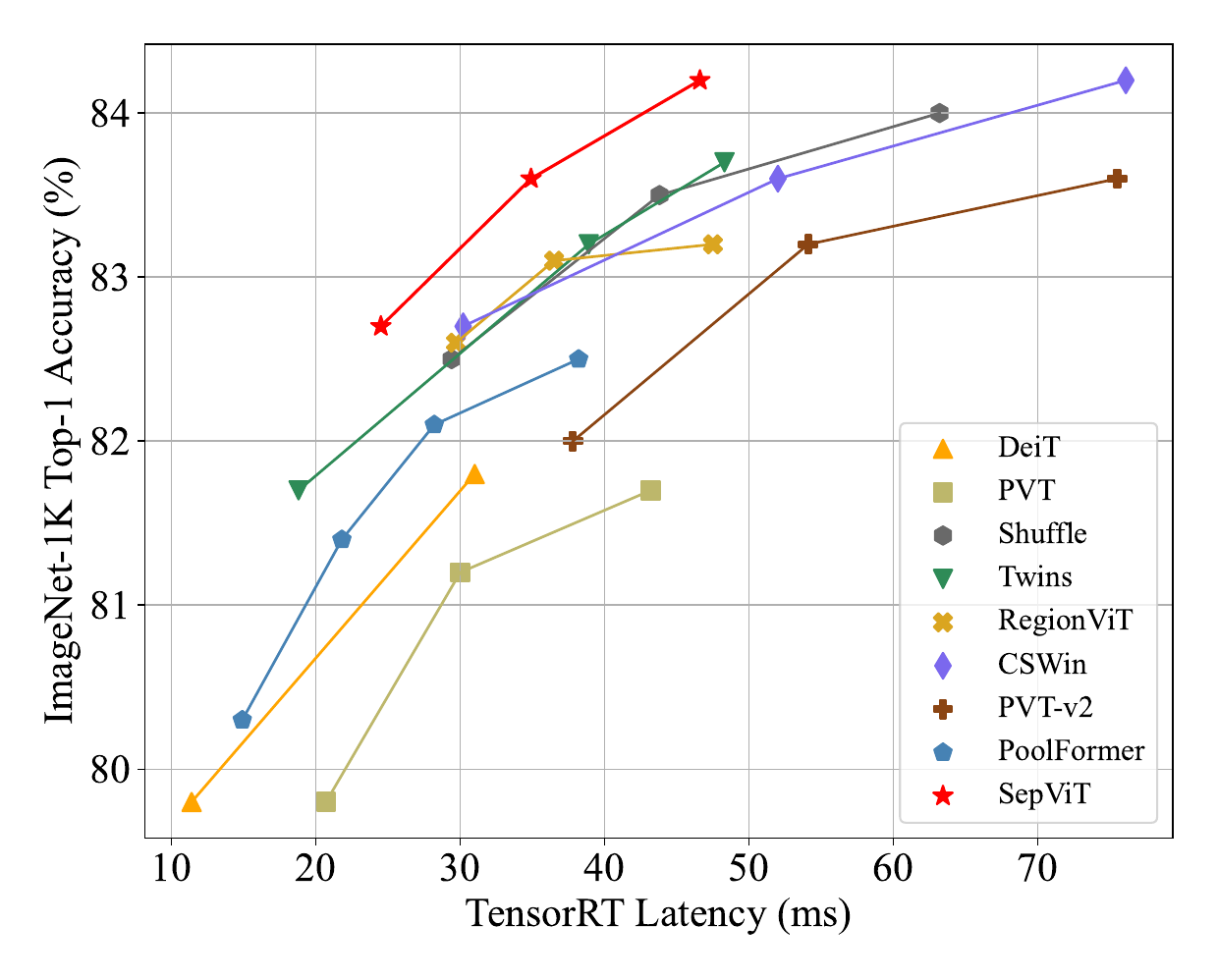}\label{throughput}}
  \caption{Comparison of throughput and latency on ImageNet-1K classification. The throughput and the latency are tested based on the PyTorch and TensorRT framework, respectively.}
  \label{fig:throughput_latency}
\end{figure*}

Primordially, ViT \cite{ViT} firstly introduces Transformer to the image recognition tasks. It splits the whole image into patches and feeds each patch as a token into Transformer. 
However, the patch-based Transformer is hard to deploy due to the computationally inefficient full-attention mechanism. 
To relieve this problem, Swin \cite{Swin} proposes the window-based self-attention to limit the computation of self-attention in non-overlapping sub-windows.
Obviously, the window-based self-attention helps to reduce the complexity to a great extent, but the shifted operator for building connection among the windows brings difficulty for ONNX or TensorRT deployment.
Twins \cite{Twins} takes advantage of the window-based self-attention and the spatial reduction attention from PVT \cite{PVT_v1} and proposes the spatially separable self-attention. 
Although Twins is deployment-friendly and achieves outstanding performance, its computational complexity is hardly reduced. 
CSWin \cite{CSWin} shows state-of-the-art performance via the novel cross-shaped window self-attention, but its throughput is low.
Albeit, with varying degrees of progress in these famous vision Transformers, most of its recent successes are accompanied with huge resource demands.

To overcome the aforementioned issues, we propose an efficient Transformer backbone, called Separable Vision Transformer (SepViT), which captures both local and global dependencies in a sequential order. A key design element of SepViT is its depthwise separable self-attention module, as shown in Fig. \ref{fig:SepViT}.
Inspired by the depthwise separable convolution in MobileNets \cite{MobileNet_v1,MobileNet_v2,Mobilenet_v3}, we re-design the self-attention module and propose the  depthwise separable self-attention, which consists of a depthwise self-attention (PSA) and a pointwise self-attention (PSA) that can correspond to depthwise and pointwise convolution in MobileNets, respectively. The depthwise self-attention is used to capture local feature within each window while the pointwise self-attention is for building connections among windows that notably improve the expressive power.
Moreover, to get the global representation of a local window, we develop a novel window token embedding, which is utilized to compute the attention relationship among windows.
Furthermore, we also extend the idea of grouped convolution from AlexNet \cite{AlexNet} to our depthwise separable self-attention and present the grouped self-attention to further improve the performance.

To demonstrate the effectiveness of SepViT, we conduct a series of experiments on some typical vision tasks, including ImageNet-1K \cite{ImageNet-1K} classification, ADE20K \cite{ADE20K} semantic segmentation, COCO \cite{COCO} object detection and instance segmentation.
The experimental results show that SepViT can achieve a better trade-off between performance and latency than other competitive vision Transformers \cite{PVT_v1,Swin,Twins,CSWin}.
As shown in Fig. \ref{fig:throughput_latency}, SepViT achieves better accuracy at the same throughput or latency constraint, and costs less inference time than the methods with the same accuracy.
Furthermore, SepViT can be expediently applied and deployed since it only contains some universal operators (e.g., transpose and matrix multiplication).
Formally, the contributions of our work can be summarized as follows:
\begin{enumerate}
  \item We design a lightweight yet efficient depthwise separable self-attention and extend it to grouped self-attention, which can achieve the information interaction within and among windows in a single Transformer block.
  \item We present the window token embedding to learn a global feature representation of each window, which is used to build the attention relationship among windows with negligible computational cost.
  \item We propose an efficient Separable Vision Transformer (SepViT), which achieves a state-of-the-art trade-off between performance and latency on various vision tasks.
\end{enumerate}

\section{Related work}
\subsection{Vision Transformer}

Vision Transformer first comes into our view when ViT \cite{ViT} is born and achieves an excellent performance on classification task.
In quick succession, a series of vision Transformers have been produced based on ViT, such as DeiT \cite{Deit}, T2T \cite{T2T}, TNT \cite{TNT}, CPVT \cite{CPVT}, etc.
Later, PVT \cite{PVT_v1} and Swin \cite{Swin} synchronously propose the hierarchical architecture which is friendly for the dense prediction tasks, such as object detection, semantic and instance segmentation.
Meanwhile, Swin \cite{Swin} as a pioneer proposes the window-based self-attention to compute attention within local windows.
Soon after, Twins \cite{Twins} and CSWin \cite{CSWin} sequentially propose the spatial separable self-attention and cross-shaped window self-attention based on the hierarchical architecture.
On the other hand, some researchers incorporate the spatial inductive biases of CNNs into Transformer. 
CoaT \cite{CoaT}, CVT \cite{CvT} and LeViT \cite{LeViT} introduce the convolutions before or after self-attentions and obtain well-pleasing results.
Regarding to the design of lightweight Transformer, Mobile-Former \cite{Mobile-Former} and MobileViT \cite{MobileViT} combine Transformer blocks with the inverted bottleneck blocks in MobileNet-V2 \cite{MobileNet_v2} in series and parallel.
Besides, another direction of research \cite{so2019evolved,Autoformer,Glit,Bossnas} is to automatically search the structure details of Transformer with neural architecture search \cite{zoph2016neural,li2022neural} technology.

\begin{figure*}[t]
    \centering
    \includegraphics[width=0.84\textwidth]{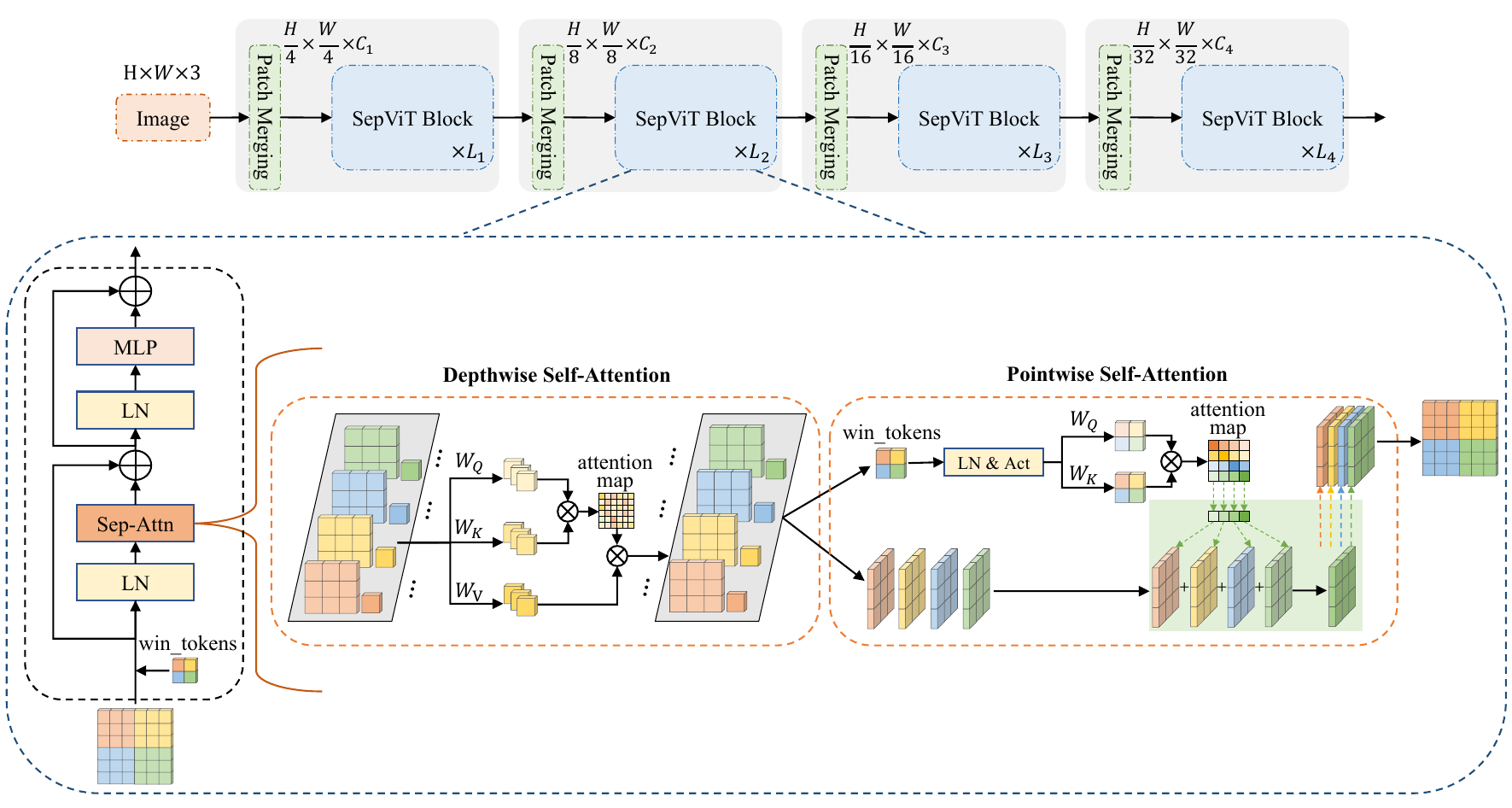}
    \caption{Separable Vision Transformer (SepViT). Top row is the overall hierarchical architecture of SepViT. Bottom row is the SepViT block and detailed visualization of our depthwise separable self-attention and the window token embedding scheme.}
    \label{fig:SepViT}
  \end{figure*}

\subsection{Lightweight Convolutions}
Many lightweight and mobile-friendly convolutions are proposed for mobile vision tasks. 
Of these, grouped convolution is the first to be proposed by AlexNet \cite{AlexNet}, which groups the feature maps and conducts the distributed training.
Then the representative work of mobile-friendly convolution must be the MobileNets \cite{MobileNet_v1,MobileNet_v2,Mobilenet_v3} with depthwise separable convolution, 
which contains a depthwise convolution for spatial information communication and a pointwise convolution for information exchange across the channels.
As time goes on, plenty of variants based on the aforementioned works are developed, such as \cite{ShuffleNet,ShuffleNet_v2,EfficientNet,GhostNet}.
In our work, we adapt the ideology of depthwise separable convolution to Transformer, which aims to reduce the Transformer's computational complexity without the sacrifice of performance.

\section{SepViT}
In this section, we first illustrate the design overview for SepViT, and then discuss some key modules within the SepViT block. Finally, we provide the architecture specifications and variants with different model sizes.

\subsection{Overview}
As illustrated in Fig. \ref{fig:SepViT}, SepViT follows the widely-used hierarchical architecture \cite{PVT_v1,Swin,Twins,CSWin} and the window-based self-attention \cite{Swin}.
Besides, SepViT also employs conditional position encoding (CPE) from \cite{CPVT,Twins}.
For each stage, there is an overlapping patch merging layer for feature map downsampling followed by a series of SepViT blocks.
The spatial resolution will be progressively reduced by 32$\times$ with either stride 4 or 2, and the channel dimension will be doubled stage by stage. 
It is worth noting that both local contextual concepts and global abstraction can be captured in a single SepViT block, while other works \cite{Swin,Twins} should employ two successive blocks to accomplish this local-global modeling.
In the SepViT block, local information communication within each window is achieved by depthwise self-attention (DSA), and global information exchange among the windows is performed via pointwise self-attention (PSA).

\subsection{Depthwise Separable Self-Attention}
\subsubsection{Depthwise Self-Attention (DSA).} 
Similar to some pioneering works \cite{Swin,Twins}, SepViT is built on top of the window-based self-attention scheme. Firstly, we perform a window partition on the input feature map. Each window can be seen as an input channel of the feature map, while different windows contain diverse information. Different from previous works, we create a window token for each window, which serves as a global representation and is used to model the attention relationship in the following pointwise self-attention module. 

Then, a depthwise self-attention (DSA) is performed on all the pixel tokens within each window as well as its corresponding window token. 
This window-wise operation is quite similar to a depthwise convolution layer in MobileNets, aiming to fuse the spatial information within each channel. 
Besides, a residual short cut is employed for DSA to prevent information loss.
The implementation of DSA can be summarized as follows:
\begin{equation}
  \label{eqution:DSA}
  \text{DSA}(z)= \text{Attention}(z \cdot W_Q, z \cdot W_K, z \cdot W_V) + z \;
\end{equation}
where $z$ is the feature tokens, consisted of the pixel tokens and window tokens.
$W_Q$, $W_K$, and $W_V$ denote three Linear layers for query, key and value computation in a regular self-attention. 
Attention means a standard self-attention operator performing as: $\text{Attention}(Q, K, V) = \text{softmax}(\frac{QK^T}{d_k})V$, in which $d_k$ denotes scaling factor. Here, it works on the each of the local window.

\subsubsection{Window Token Embedding.}
To establish the attention relationship among windows, a straightforward solution is to employ all pixel tokens of each window. However, it will bring huge computational costs and lead the whole model to be complex.
To alleviate this issue, we present a window token embedding scheme, which leverages a single token to encapsulate the core information for each sub-window. 
This window token can be initialized either as a fixed zero vector or a learnable vector. While passing through DSA, there is an informational interaction between the window token and pixel tokens in each window. Thus the window token can learn a global representation of this window. 
Thanks to the effective window token, we can build the attention relationship among windows with negligible computational cost.

\subsubsection{Pointwise Self-Attention (PSA).} 
The famous pointwise convolution in MobileNets is utilized to fuse the information from different channels. 
In our work, we imitate pointwise convolution to develop the pointwise self-attention (PSA) module to establish connections among windows. 
PSA is mainly used to fuse the information across windows and obtain a final representation of the input feature map as well. 
More specifically, we extract the feature maps and window tokens from the output of DSA. Then, window tokens are used to compute the attention relationship among windows and generate the attention map after a LayerNormalization (LN) layer and a GELU activation function (Act). 
Meanwhile, we directly treat the feature maps as the value branch of PSA without any other extra operation. 

With the attention map and the feature maps which are in the form of windows, we perform an attention computation among the windows for global information exchange. 
PSA adopts a residual connection as well. Formally, the implementation of PSA can be depicted as follows:
\begin{equation}
  \label{eqution:PSA}
  \begin{aligned}
    \text{PSA}(z, wt) = \text{At}&\text{tention}(\text{GELU}(\text{LN}(wt)) \cdot W_Q, \; \\
                        & \text{GELU}(\text{LN}(wt)) \cdot W_K, z) + z \; 
  \end{aligned}
\end{equation}
where $wt$ denotes the window token. Here, Attention is a standard self-attention operator as well, but works on the whole feature maps that are in the form of windows $z$.

\subsubsection{Complexity Analysis.}
\label{sec:complexity}
Given an input feature with size $H \times W \times C$, the computational complexity of the multi-head self-attention (MSA) is $4HWC^2 + 2H^2W^2C$ in the global Transformer block of ViT \cite{ViT}.
The complexity of the MSA in a window-based Transformer with window size $M \times M$ (Usually, $M$ is a common factor of $H$ and $W$, so the number of the windows is $N=\frac{HW}{M^2}$) can be decreased to $4HWC^2 + 2M^2HWC$ in Swin \cite{Swin}.
As for the depthwise separable self-attention in SepViT, the complexity contains two parts, DSA and PSA.

{\bf DSA.} Building on top of a window-based self-attention, DSA shares a similar computational cost to it. Additionally, the introduction of window tokens will cause an extra cost, but it is negligible compared to the overall cost of DSA. The complexity of DSA can be calculated as follows:
\begin{equation}
  \varOmega(\text{DSA}) = 3HWC^2 + 3NC^2 + 2N(M^2+1)^2C \; 
\end{equation}
where $3NC^2$ is the extra cost of encoding window tokens in Linear layers.
$2N(M^2+1)^2C$ indicates the matrix multiplications involved in the self-attention for $N$ windows,  
where $M^2+1$ represents the $M^2$ pixel tokens in a window and its corresponding window token.
Thanks to that the number of sub-windows $N$ is usually a small value, the extra costs caused by window tokens can be ignored.

{\bf PSA.} Since the window token summarizes the global information of a local window, the proposed PSA helps to efficiently perform information exchange among windows in a window level instead of pixel level. Concretely, the complexity of PSA is given as follows:
\begin{equation}
  \varOmega(\text{PSA}) = HWC^2 + 2NC^2 + N^2C + NHWC \; 
\end{equation}
where $2NC^2$ represents the little cost of computing the query and key with $N$ window tokens, while the value branch is zero-cost. $N^2C$ indicates the computation of generating attention map with $N$ window tokens in a window level, which saves computational cost for PSA to a great extent. Finally, $NHWC$ indicates the matrix multiplication between the attention map and the feature maps.

\subsection{Grouped Self-Attention}
\begin{figure}
  \centering
  \includegraphics[height=4cm]{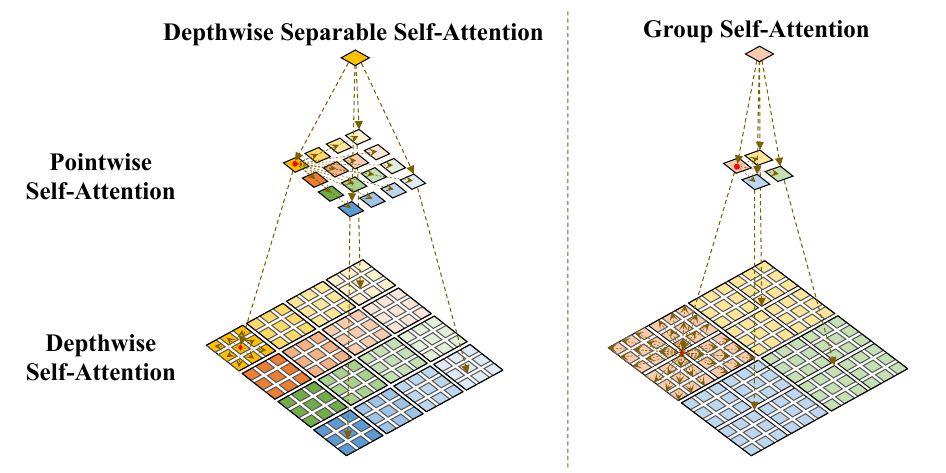}
  \caption{A macro view of the similarities and differences between the depthwise separable self-attention and the grouped self-attention.}
  \label{fig:grouped_attn}
\end{figure}

Motivated by the excellent performance of grouped convolution in visual recognition \cite{AlexNet}, we make an extension of our depthwise separable self-attention and propose the grouped self-attention.
As shown in Fig. \ref{fig:grouped_attn}, we splice several neighboring sub-windows to form a larger window, which is similar to dividing the windows into groups and conducting a depthwise self-attention communication inside a group of windows. In this way, the grouped self-attention can capture long-range visual dependencies of multiple windows.
In terms of computational cost and performance gains, the grouped self-attention has a certain extra cost compared with the depthwise separable self-attention but results in a better performance.
Ultimately, we apply the block with grouped self-attention (GSA) to SepViT and run it alternately with the depthwise separable self-attention block (DSSA) in the late stages of the network.

\subsection{SepViT Block}
Formally, our SepViT block can be formulated as follows:
\begin{align}
 \label{eqution:SepViT}
 & \tilde{z}^l = \text{Concat}(z^{l-1}, wt) \; \\
 & \ddot{z}^l = \text{DSA}(\text{LN}(\tilde{z}^l)) \; \\
 & \dot{z}^l, \dot{wt} = \text{Slice}(\ddot{z}^l) \; \\
 & \hat{z}^l = \text{PSA}(\dot{z}^l, \dot{wt}) + z^{l-1} \; \\
 & z^l = \text{MLP}(\text{LN}(\hat{z}^l)) + \hat{z}^l \;
\end{align}
where $\ddot{z}^l$, $\hat{z}^l$ and $z^{l}$ denote the outputs of the DSA, PSA and SepViT block $l$, respectively.
$\dot{z}^l$ and $\dot{wt}$ are feature maps and the learned window tokens.
Concat represents the concatenation operation while Slice represents the slice operation.

\begin{figure}[t]
  \centering
  \includegraphics[height=4cm]{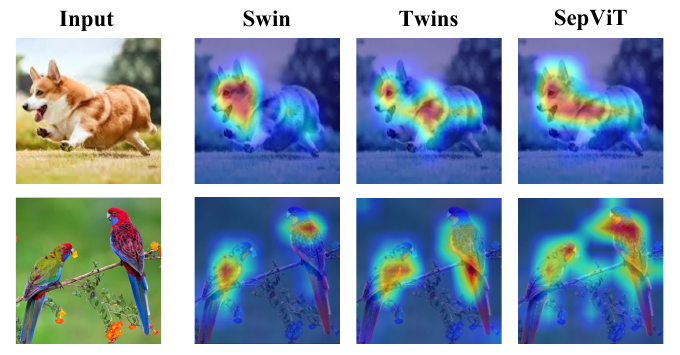}
  \caption{Grad-CAM \cite{gradcam} activation maps of Swin-T, Twins-SVT-S and SepViT-T. SepViT shows a prominent capability of capturing a wider range information.}
  \label{fig:visualizaiton}
\end{figure}

{\bf Comparison of Complexity and Inference Speed.} 
We compare SepViT block (DSSA) with two other SOTA blocks (Swin \cite{Swin}, Twins \cite{Twins}). As we stated before, the information interaction within and among windows is completed in a single SepViT block, while Swin \cite{Swin} and Twins \cite{Twins} require two successive blocks. 
As shown in Table \ref{tab:inference_speed}, we can observe that the SepViT block only costs about half the MACs of its competitors. The reason lies in two aspects. Firstly, SepViT block is more lightweight. Secondly, SepViT block removes many redundant layers, e.g., there is only one MLP layer and two LN layers in a single SepViT block while there are double MLP and LN layers in two successive Swin or Twins blocks.
On the other hand, we make a comparison of inference speed between a single SepViT blcok with these two-blcok pattern works (Swin and Twins). Table \ref{tab:inference_speed} reports that SepViT blcok is about 60\% faster on Pytorch and 55\% faster on TensorRT.

\begin{table}[ht]
  \centering
  \resizebox{0.45\textwidth}{!}{
  \begin{tabular}{@{}c|ccc@{}}
  \toprule 
  Block Type    & FLOPs(G) & Throughput(img/s)   & Latency(ms)    \\  \midrule
  Swin          & 0.77      & 1346                   & -              \\
  Twins         & 0.78      & 1403                   & 5.3            \\      
  {\bf SepViT}  & 0.40      & {\bf 2146}             & {\bf 2.4}      \\      
  \bottomrule
  \end{tabular}}
  \caption{Comparison of complexity and inference speed between SepViT block and the two-blcok pattern works.}
  \label{tab:inference_speed}
\end{table}

\section{Experimental Results}
To present a fair comparison with other ViTs, we mainly propose the SepViT-T/S/B variants. A series of experiments are conducted with them to verify the capability of SepViT, including classification, segmentation and detection.
Architecture configurations and implementation details are provided in Appendix.

\begin{table}[t]
  \centering
  \scalebox{0.59}{
  \begin{tabular}{c|cccc|c}
  \toprule
  \multirow{2}{*}{Method}             & Param         & FLOPs          & Throughput     & Latency        & Top-1 Acc  \\ 
                                      & (M)           & (G)            & (img/s)     & (ms)           & (\%)       \\   \midrule
  \multicolumn{6}{c}{ConvNet}                                                                                         \\   \midrule
  ConvNeXt-T\cite{ConvNext}           & 29.0          & 4.5            & -              & 19.0           & 82.1       \\ 
  ConvNeXt-S\cite{ConvNext}           & 50.0          & 8.7            & -              & 28.1           & 83.1       \\ 
  ConvNeXt-B\cite{ConvNext}           & 88.0          & 15.4           & -              & 37.3           & 83.9       \\ \midrule
  
  \multicolumn{6}{c}{Transformer}                                                                                      \\   \midrule
  DeiT-Small/16\cite{Deit}            & 22.0          & 4.6            & 406            & 11.4           & 79.9        \\
  T2T-ViT-14\cite{T2T}                & 22.0          & 5.2            & -              & -              & 81.5        \\
  TNT-S\cite{TNT}                     & 23.8          & 5.2            & -              & -              & 81.3        \\
  CoaT-Lite-Small\cite{CoaT}          & 20.0          & 4.0            & -              & -              & 81.9        \\
  CvT-13\cite{CvT}                    & 20.0          & 4.5            & -              & -              & 81.6        \\
  PVT-Small\cite{PVT_v1}              & 24.5          & 3.8            & 794            & 20.7           & 79.8        \\
  Swin-T\cite{Swin}                   & 29.0          & 4.5            & 704            & -              & 81.3        \\
  Shuffle-T\cite{huang2021shuffle}    & 29.0          & 4.6            & 712            & 29.4           & 82.5        \\ 
  Twins-S\cite{Twins}                 & 24.0          & 2.9            & 979            & 18.8           & 81.7        \\ 
  RegionViT-S\cite{regionvit}         & 40.6          & 5.3            & 671            & 29.7           & 82.6        \\
  CSWin-T\cite{CSWin}                 & 23.0          & 4.3            & 627            & 30.2           & 82.7        \\
  PVT-v2-B2\cite{PVT_v2}              & 25.4          & 4.0            & 664            & 37.8           & 82.0        \\
  PoolFormer-S24\cite{metaformer}     & 21.0          & 3.5            & 993            & 14.9           & 80.3        \\ 
  PoolFormer-S36\cite{metaformer}     & 31.0          & 5.1            & 764            & 21.8           & 81.4        \\ 
  {\bf SepViT-T(ours)}                & {\bf 31.2}    & {\bf 4.5}      & {\bf 729}      & {\bf 24.5}     & {\bf 82.7}  \\  \midrule                            
  
  T2T-ViT-19\cite{T2T}                & 39.2          & 8.9            & -              & -              & 81.9        \\
  CoaT-Lite-Medium\cite{CoaT}         & 45.0          & 9.8            & -              & -              & 83.6        \\
  CvT-21\cite{CvT}                    & 32.0          & 7.1            & -              & -              & 82.5        \\
  CMT-S\cite{Cmt}                    & 25.1          & 4.0            & -              & 52.0              & 83.5        \\
  PVT-Medium\cite{PVT_v1}             & 44.2          & 6.7            & 511            & 30.0           & 81.2        \\
  Swin-S\cite{Swin}                   & 50.0          & 8.7            & 412            & -              & 83.0        \\
  Shuffle-S\cite{huang2021shuffle}    & 50.0          & 8.9            & 417            & 43.8           & 83.5        \\ 
  Twins-B\cite{Twins}                 & 56.0          & 8.6            & 433            & 38.9           & 83.2        \\
  RegionViT-M\cite{regionvit}         & 41.2          & 7.4            & 448            & 36.5           & 83.1        \\
  CSWin-S\cite{CSWin}                 & 35.0          & 6.9            & 390            & 52.0           & 83.6        \\
  PVT-v2-B3\cite{PVT_v2}              & 45.2          & 6.9            & 443            & 54.1           & 83.2        \\
  PoolFormer-M36\cite{metaformer}     & 56.0          & 9.0            & 494            & 28.2           & 82.1        \\ 
  {\bf SepViT-S(ours)}                & {\bf 46.6}    & {\bf 7.5}      & {\bf 471}      & {\bf 34.9}     & {\bf 83.6}  \\  \midrule                      
  
  DeiT-Base/16\cite{Deit}             & 86.6          & 17.6           & 273            & 31.0              & 81.8        \\
  T2T-ViT-24\cite{T2T}                & 64.1          & 14.1           & -              & -              & 82.3        \\
  TNT-B\cite{TNT}                     & 66.0          & 14.1           & -              & -              & 82.8        \\
  PVT-Large\cite{PVT_v1}              & 61.4          & 9.8            & 357            & 43.2           & 81.7        \\
  Swin-B\cite{Swin}                   & 88.0          & 15.4           & 255            & -              & 83.3        \\
  Shuffle-B\cite{huang2021shuffle}    & 88.0          & 15.6           & 259            & 63.2           & 84.0        \\ 
  Twins-L\cite{Twins}                 & 99.2          & 15.1           & 271            & 48.3           & 83.7        \\
  RegionViT-B\cite{regionvit}         & 72.7          & 13.0           & 286            & 47.5           & 83.2        \\
  CSWin-B\cite{CSWin}                 & 78.0          & 15.0           & 216            & 76.1           & 84.2        \\
  PVT-v2-B4\cite{PVT_v2}              & 62.6          & 10.1           & 298            & 75.5           & 83.6        \\
  PoolFormer-M48\cite{metaformer}     & 73.0          & 11.8           & 337            & 38.2           & 82.5        \\ 
  {\bf SepViT-B(ours)}                & {\bf 82.3}    & {\bf 13.1}     & {\bf 302}      & {\bf 46.6}     & {\bf 84.2}  \\  
  \bottomrule
  \end{tabular}}
  \caption{Comparison of different state-of-the-art methods on ImageNet-1K classification. 
  Throughput and latency are tested based on the PyTorch framework with a V100 GPU (batch size=192) and TensorRT framework with a T4 GPU (batch size=8).}
  \label{tab:ImageNet_1K}
\end{table}

\subsection{ImageNet-1K Classification}

We first carry out the image classification experiment on the ImageNet-1K \cite{ImageNet-1K}.
As the results shown in Table \ref{tab:ImageNet_1K}, compared to the latest state-of-the-art vision Transformers, SepViT achieves the best trade-off between accuracy and latency.
To be more specific, compared with these famous ViTs (e.g. Swin), SepViT-S/B achieve 83.6\% and 84.2\% top-1 accuracy surpassing Swin-S/B by 0.6\% and 0.9\% with about 14\% fewer FLOPs and 15\% faster inference speed.
As for the tiny variant, SepViT-T outperforms the Swin-T by 1.4\% with the same FLOPs of 4.5G and a faster throughput as well. 
Meanwhile, SepViT-T/S/B outperforms RegionViT-S/M/B by 1.1\%, 0.5\% and 1.0\% under a similar TensorRT latency.
On the other hand, compared with the recent SOTA methods with the similar accuracy (e.g. CSWin and Shuffle Transformer), 
both of our small and base variants cost about 40\% less inference latency on TensorRT than CSWin-S/B, and SepViT-T is 19\% faster than CSWin-T as well. 
Compared to Shuffle Transformer, SepViT-T/S/B achieve a better performance than Shuffle-T/S/B while being 17\%, 20\% and 27\% faster on TensorRT.
Moreover, SepViT also shows very a promising performance by comparing with the CNNs counterparts under the similar FLOPs (e.g. RegNetY and ConvNeXt).

\begin{table*}[ht]
  \centering
  \resizebox{0.9\textwidth}{!}{
  \begin{tabular}{c|c|ccc|ccc}
  \toprule
  \multirow{2.5}{*}{Backbone}     & \multirow{2.5}{*}{\shortstack{Latency \\ (ms)}}   & \multicolumn{3}{c|}{Semantic FPN 80k}   & \multicolumn{3}{c}{UperNet 160k}          \\ \cmidrule(l){3-8}
                                              &               & Param(M)    & FLOPs(G)   & mIoU(\%)      & Param(M)  &   FLOPs(G)     & mIoU/MS mIoU(\%)        \\   \midrule
  ResNet50\cite{ResNet}                       & 18.8          & 28.5        & 183        & 36.7          & -            & -           & -/-                     \\
  ConvNeXt-T\cite{ConvNext}                   & 78.5          & -           & -          & -             & 60.0         & 939         & -/46.7                  \\
  PVT-Small\cite{PVT_v1}                      & 129.6         & 28.2        & 161        & 39.8          & -            & -           & -/-                     \\
  Swin-T\cite{Swin}                           & -             & 31.9        & 182        & 41.5          & 59.9         & 945         & 44.5/45.8               \\
  Shuffle-T\cite{huang2021shuffle}            & 168.9         & -           & -          & -             & 60.0         & -           & 46.6/47.6               \\
  Twins-S\cite{Twins}                         & 127.2         & 28.3        & 144        & 43.2          & 54.4         & 901         & 46.2/47.1               \\
  PoolFormer-S36\cite{metaformer}             & 87.7          & 34.6        & -          & 42.0          & -            & -           & -/-                     \\
  {\bf SepViT-T(ours)}                        & {\bf 139.7}   & {\bf 38.8}  & {\bf 181}  & {\bf 44.8}    & {\bf 66.8}   & {\bf 940}   & {\bf 47.4}/{\bf 48.3}   \\ \midrule
  
  ResNet101\cite{ResNet}                      & 32.8          & 47.5        & 260        & 38.8          & 86.0         & 1092        & -/44.9                  \\
  ConvNeXt-S\cite{ConvNext}                   & 131.5         & -           & -          & -             & 82.0         & 1027        & -/49.6                  \\
  PVT-Medium\cite{PVT_v1}                     & 184.2         & 48.0        & 219        & 41.6          & -            & -           & -/-                     \\
  Swin-S\cite{Swin}                           & -             & 53.2        & 274        & 45.2          & 81.3         & 1038        & 47.6/49.5               \\
  Shuffle-S\cite{huang2021shuffle}            & 283.7         & -           & -          & -             & 81.0         & -           & 48.4/49.6               \\
  Twins-B\cite{Twins}                         & 231.8         & 60.4        & 261        & 45.3          & 88.5         & 1020        & 47.7/48.9               \\
  PoolFormer-M36\cite{metaformer}             & 127.8         & 59.8        & -          & 42.4          & -            & -           & -/-                     \\
  {\bf SepViT-S(ours)}                        & {\bf 201.5}   & {\bf 55.4}  & {\bf 244}  & {\bf 46.6}    & {\bf 83.4}   & {\bf 1003}  & {\bf 48.6}/{\bf 49.9}   \\   \midrule
  
  ResNeXt101-64$\times$4d\cite{ResNeXt}       & 65.7          & 86.4        & -          & 40.2          & -            & -           & -/-                     \\
  ConvNeXt-B\cite{ConvNext}                   & 181.3         & -           & -          & -             & 122.0        & 1170        & -/49.9                  \\
  PVT-Large\cite{PVT_v1}                      & 359.3         & 65.1        & 283        & 42.1          & -            & -           & -/-                     \\
  Swin-B\cite{Swin}                           & -             & 91.2        & 422        & 46.0          & 121.0        & 1188        & 48.1/49.7               \\
  Shuffle-B\cite{huang2021shuffle}            & 521.2         & -           & -          & -             & 121.0        & -           & 49.0/50.5               \\
  Twins-L\cite{Twins}                         & 409.7         & 103.7       & 404        & 46.7          & 133.0        & 1164        & 48.8/49.7               \\
  PoolFormer-M48\cite{metaformer}             & 168.7         & 77.1        & -          & 42.7          & -            & -           & -/-                     \\
  {\bf SepViT-B(ours)}                        & {\bf 366.3}   & {\bf 94.7}  & {\bf 367}  & {\bf 47.6}    & {\bf 124.8}  & {\bf 1128}  & {\bf 49.6}/{\bf 51.0}   \\ 
  \bottomrule
  \end{tabular}}
  \caption{Comparison of different backbones on ADE20K semantic segmentation task. Latency is the TensorRT runtime of the backbone with the input size of 512$\times$512 (batch size=8). FLOPs are measured with the input size of 512$\times$2048.}
  \label{tab:ADE20K}
\end{table*}
\subsection{Semantic Segmentation}
To further verify the capacity of SepViT, we conduct the semantic segmentation experiment on ADE20K \cite{ADE20K}. In Table \ref{tab:ADE20K}, we make a comparison with the recent vision Transformer and CNN backbones.
Based on the Semantic FPN framework \cite{Semantic_FPN}, SepViT-T/S/B surpasses Swin-T/S/B by 3.3\%, 1.4\% and 1.6\% mIoU with fewer FLOPs, and outperforms Twins-S/B/L by 1.6\%, 1.3\% and 0.9\% mIoU with faster TensorRT inference time. Meanwhile, SepViT shows great advantage over CNNs. SepViT-T/S gain a promising performance of 8.1\% and 7.8\% mIoU than ResNet50 and ResNet101.
By contrast with Swin on the UperNet framework \cite{UperNet}, SepViT-T/S/B achieves 2.9\%, 1.0\%, 1.5\% higher mIoU on single-scale testing and 2.5\%, 0.4\%, 1.3\% higher mIoU in terms of multi-scale (MS) testing. 
Compared to the famous ConvNeXt, SepViT-T/S/B achieve 1.6\%, 0.3\% and 1.1\% higher mIoU on multi-scale testing. Extensive experiments reveal that SepViT shows great potential on segmentation tasks.

\subsection{Object Detection and Instance Segmentation}

Next, we evaluate SepViT on the objection detection and instance segmentation tasks with RetinaNet \cite{RetinaNet} and Mask R-CNN \cite{Mask_RCNN} frameworks on COCO \cite{COCO}.
The experimental results in Table \ref{tab:COCO} show that SepViT achieves a state-of-the-art performance compared with the recent vision Transformers and some famous CNNs, and its TensorRT runtime is competitive as well.
Based on the RetinaNet framework, SepViT-T/S surpass Swin-T/S by 3.3, 1.9 AP with fewer FLOPs in the 1$\times$ experiment.
For the 3$\times$ schedule, SepViT-S outperforms Twins-B and RegionViT-B by 1.0 and 1.8 AP with a faster inference speed at the same time.
As for the Mask R-CNN framework, SepViT-T outperforms Swin-T by 3.3 AP$^b$, 2.8 AP$^m$ with 1$\times$ schedule, and 2.0 AP$^b$, 2.2 AP$^m$ with 3$\times$ schedule. For the SepViT-S variant, it achieves a similar performance gain while saving a certain amount of computation overhead and inference latency.
Furthermore, compared with the famous CNNs in the 3$\times$ experiment, SepViT-T surpasses ConvNeXt-T by 1.8 AP$^b$, 2.1 AP$^m$, and SepViT-S outperforms ResNet101 by 6.6 AP$^b$, 6.1 AP$^m$.
Thus, experimental results on these object detection and instance segmentation tasks demonstrate SepViT's capability of getting better trade-off between performance and latency.

\begin{table*}[t]
  \centering
  \resizebox{0.9\textwidth}{!}{
  \begin{tabular}{c|c|cc|cc|cc|cccc}
  \toprule
  \multirow{3}{*}{Backbone}               & \multirow{3}{*}{\shortstack{Latency \\ (ms)}}  & \multicolumn{4}{c|}{RetinaNet}            & \multicolumn{6}{c}{Mask R-CNN}       \\ \cline{3-12} 
                                          &             & Params      & FLOPs     & 1$\times$     & 3$\times$ + MS      & Params      & FLOPs         & \multicolumn{2}{c}{1$\times$}           & \multicolumn{2}{c}{3$\times$ + MS} \\ \cline{5-6} \cline{9-12}
                                          &             & (M)         & (G)       & AP            & AP                  & (M)         & (G)           & AP$^b$         & AP$^m$          & AP$^b$        & AP$^m$            \\  \midrule
  ResNet50\cite{ResNet}                   & 29.7        & 37.7        & 239       & 36.3          & 39.0                & 44.2        & 260           & 38.0           & 34.4            & 41.0          & 37.1              \\                                                                                                                                                                  
  ConvNeXt-T\cite{ConvNext}               & 135.2       & -           & -         & -             & -                   & -           & 262           & -              & -               & 46.2          & 41.7              \\
  PVT-Small\cite{PVT_v1}                  & 215.7       & 34.2        & 226       & 40.4          & 42.2                & 44.1        & 245           & 40.4           & 37.8            & 43.0          & 39.9              \\                                                                                                                                                                      
  Swin-T\cite{Swin}                       & -           & 38.5        & 245       & 41.5          & 43.9                & 47.8        & 264           & 42.2           & 39.1            & 46.0          & 41.6              \\                                                                                                                                                                          
  Shuffle-T\cite{huang2021shuffle}        & 291.4       & -           & -         &  -            & -                   & 48.0        & 268           & -              & -               & 46.8          & 42.3              \\
  Twins-S\cite{Twins}                     & 205.9       & 34.4        & 210       & 43.0          & 45.6                & 44.0        & 228           & 43.4           & 40.3            & 46.8          & 42.6              \\                                                                                                                                                                      
  RegionViT-S\cite{regionvit}             & 287.9       & 40.8        & 193       & 42.2          & 45.8                & 50.1        & 171           & 42.5           & 39.5            & 46.3          & 42.3              \\
  PVTv2-B2\cite{PVT_v2}                   & 387.7       & 35.1        & 231       & 44.6          & -                   & 45.0        & -             & 45.3           & 41.2            & -             & -                 \\                                                                                                                                                                    
  PoolFormer-S36\cite{metaformer}         & 142.1       & 40.6        & -         & 39.5          & -                   & 50.5        & -             & 41.0           & 37.7            & -             & -                 \\
  {\bf SepViT-T(ours)}                    & {\bf 241.8} & {\bf 45.4}  & {\bf 243} & {\bf 44.8}    & {\bf 46.8}          & {\bf 54.7}  & {\bf 261}     & {\bf 45.5}     & {\bf 41.9}      & {\bf 48.0}    & {\bf 43.8}        \\  \midrule                                                                                                                                                                     
  
  ResNet101\cite{ResNet}                  & 45.9        & 58.0        & 315       & 38.5          & 40.9                & 63.2        & 336           & 40.4           & 36.4            & 42.8          & 38.5              \\                                                                                                                                                                    
  ResNeXt101-32$\times$4d\cite{ResNeXt}   & 105.1       & 56.4        & 319       & 39.9          & 41.4                & 62.8        & 340           & 41.9           & 37.5            & 44.0          & 39.2              \\ 
  PVT-Medium\cite{PVT_v1}                 & 341.2       & 53.9        & 283       & 41.9          & 43.2                & 63.9        & 302           & 42.0           & 39.0            & 44.2          & 40.5              \\                                                                                                                                                                          
  Swin-S\cite{Swin}                       & -           & 59.8        & 335       & 44.5          & 46.3                & 69.1        & 354           & 44.8           & 40.9            & 48.5          & 43.3              \\                                                                                                                                                                    
  Shuffle-S\cite{huang2021shuffle}        & 408.2       & -           & -         &  -            & -                   & 69.0        & 359           & -              & -               & 48.4          & 43.3              \\
  Twins-B\cite{Twins}                     & 398.8       & 67.0        & 326       & 45.3          & 46.9                & 76.3        & 340           & 45.2           & 41.5            & 48.0          & 43.0              \\                                                                                                                                                                  
  RegionViT-B\cite{regionvit}             & 451.4       & 83.4        & 309       & 43.3          & 46.1                & 92.2        & 288           & 43.5           & 40.1            & 47.2          & 43.0              \\
  PVTv2-B3\cite{PVT_v2}                   & 439.1       & 55.0        & 288       & 45.9          & -                   & 64.9        & -             & 47.0           & 42.5            & -             & -                 \\                                                                                                                                                                  
  {\bf SepViT-S(ours)}                    & {\bf 375.8} & {\bf 61.9}  & {\bf 302} & {\bf 46.4}    & {\bf 47.9}          & {\bf 71.3}  & {\bf 321}     & {\bf 47.2}     & {\bf 42.9}      & {\bf 49.4}    & {\bf 44.6}        \\                                                                                                                                                                     
  \bottomrule
  \end{tabular}}
  \caption{Comparison of different backbones on object detection and instance segmentation tasks with RetinaNet and Mask R-CNN frameworks. 
  Latency is the TensorRT runtime of the backbone with the input size of 800$\times$800 (batch size=8). FLOPs are measured with the inpus size of $800 \times 1280$.
  The superscript $b$ and $m$ denote the box detection and mask instance segmentation.}
  \label{tab:COCO}
\end{table*}

\begin{table*}[ht]
  \centering
  \resizebox{0.9\textwidth}{!}{
  \begin{tabular}{@{}c|ccc|cccc|c@{}}
  \toprule
  Model                       & DSSA            & GSA         & LWT       & Param(M)    & FLOPs(G)    & Throughput(img/s)  & Latency(ms)     & Top-1 Acc(\(\% \))       \\  \midrule
  Swin-T+CPVT \cite{Twins}    &                 &             &           & 28.0        & 4.4         & 704                   & -               & 81.2            \\ \midrule 
  \multirow{3}{*}{SepViT-T}   & $\surd$         &             &           & 32.1        & 4.4         & 740                   & 23.1            & 82.2            \\    
                              & $\surd$         & $\surd$     &           & 31.2        & 4.5         & 729                   & 24.5            & 82.5            \\   
                              & $\surd$         & $\surd$     & $\surd$   & 31.2        & 4.5         & 729                   & 24.5            & {\bf82.7}       \\ 
  \bottomrule
  \end{tabular}}
  \caption{Ablation studies of the key components in our SepViT. LWT means the learnable window tokens.}
  \label{tab:Ablation1}
\end{table*}
\subsection{Ablation Study}
To better demonstrate the significance of each key component, including depthwise separable self-attention, grouped self-attention and window token embedding scheme in SepViT, we conduct a series of ablation experiments on ImageNet-1K classification with SepViT-T variant.

\subsubsection{Efficient Components.}
As mentioned above, SepViT adopts the conditional position encoding (CPE) \cite{CPVT}. Therefore, we take the Swin-T+CPVT reported in \cite{Twins} as the baseline.
As shown in Table \ref{tab:Ablation1} where each component is added in turn to verify their benefits, SepViT-T simply equipped with the depthwise separable self-attention block (DSSA) outperforms Swin+CPVT by 1.0\% (from 81.2\% to 82.2\%) and it is much faster with the throughput of 740 images/s.
After employing grouped self-attention block (GSA) and DSSA alternately in the second and third stages, we gain an accuracy improvement of 0.3\% (from 82.2\% to 82.5\%).
Moreover, we study whether it makes a difference if the window token is initialized with a fixed zero vector or a learnable vector.
In contrast to the fixed zero initialization scheme, the learnable window token (LWT) helps SepViT-T to boost the performance to 82.7\%.
\begin{table}
  \centering
  \resizebox{0.48\textwidth}{!}{
  \begin{tabular}{@{}c|ccc|c@{}}
  \toprule
  Method            & Param(M)    & FLOPs(G)    & Throughput(img/s)      & Top-1 Acc(\(\% \)) \\  \midrule
  {\bf Win\_Tokens} & {\bf 31.2}  & {\bf 4.5}   & {\bf 729}                 & {\bf 82.7}          \\
  Avg\_Pooling      & 31.2        & 4.5         & 734                       & 82.2          \\
  Dw\_Conv          & 31.3        & 4.5         & 720                       & 82.3            \\
  \bottomrule
  \end{tabular}}
  \caption{Comparison of different approaches of getting the global representation of each window in SepViT.}
  \label{tab:Ablation2}
\end{table}

\subsubsection{Window Token Embedding.}
To verify the effectiveness of learning the global representation of each window with our window token embedding scheme (Win\_Tokens), we further study some other methods that directly get the global representations from the output feature maps of DSA, such as average pooling (Avg\_Pooling) and depthwise convolution (DW\_Conv). 
As the results illustrated in Table \ref{tab:Ablation2}, our window token embedding scheme achieves the best performance among these approaches. Meanwhile, the comparison of parameters and FLOPs between Win\_Token and Avg\_Pooling methods demonstrates that our window token embedding scheme costs negligible computation.

\subsubsection{Combination Types.} 
We have studied the combinations of DSSA and GSA in SepViT. 
Particularly, we just apply GSA in the last few stages due to the high computation and resolution in stage 1.
As the result shows in Table \ref{tab:Ablation3}, the best performance of 82.7\% is achieved by putting DSSA before GSA and running them alternately in stage 2 and 3.

\begin{table}
  \centering
  \resizebox{0.48\textwidth}{!}{
  \begin{tabular}{@{}c|ccc|c@{}}
  \toprule
  Block Type          & Param(M)    & FLOPs(G)    & Throughput(img/s)      & Top-1 Acc(\%)       \\  \midrule
  (D, D, D, D)              & 32.1        & 4.4         & 740                       & 82.2            \\   
  (D, D, D\&G, D)           & 31.2        & 4.5         & 733                       & 82.4            \\
  {\bf (D, D\&G, D\&G, D)}  & {\bf 31.2}  & {\bf 4.5}   & {\bf 729}                 & {\bf 82.7}            \\
  (D, G\&D, G\&D, D)        & 31.2        & 4.5         & 729                       & 82.3            \\
  \bottomrule
  \end{tabular}}
  \caption{Comparision of different combination types of DSSA (D) and GSA (G) in SepViT.}
  \label{tab:Ablation3}
\end{table}

\section{Conclusion}
In this paper, we have presented an efficient Separable Vision Transformer, dubbed SepViT, which consists of three core designs. 
Firstly, depthwise separable self-attention enables SepViT to achieve information interaction within and among the windows in a single block.
Next, window token embedding scheme helps SepViT to build the attention relationship among windows with negligible computational cost.
Thirdly, the extended grouped self-attention enables SepViT to capture long-range visual dependencies across multiple windows for better performance.
Experimental results on various vision tasks verify that SepViT achieves a state-of-the-art trade-off between performance and latency.

\bibliography{aaai23}

\newpage
\section{Appendix}
\subsection{Architecture Configurations}
For fair comparison, we propose the SepViT-T/S/B variants.
Moreover, we also design the SepViT-Lite variant with a very light model size. 
The specific configurations of SepViT variants are listed in Table \ref{tab:configs}, the block depth of SepViT is smaller in some stages than the competitors since SepViT is more efficient. DSSA and GSA denote the blocks with depthwise separable self-attention and grouped self-attention, respectively. Additionally, the expansion ratio of each MLP layer is set as 4, the window sizes are 7$\times$7 and 14$\times$14 for DSSA and GSA in all SepViT variants.

\begin{table}[h]
  \centering
  \resizebox{0.48\textwidth}{!}{
  \begin{tabular}{@{}c|c|c|c|c@{}}
  \toprule
  Configs   & SepViT-Lite           & SepViT-T              & SepViT-S              & SepViT-B                  \\ \midrule
  Blocks          & [1, 2, 6, 2]          & [1, 2, 6, 2]          & [1,  2, 14, 2]        & [1,  2, 14, 2]            \\ \midrule
  Channels        & [32, 64, 128, 256]    & [96, 192, 384, 768]   & [96, 192, 384, 768]   & [128, 256, 512, 1024]     \\ \midrule
  Heads           & [1, 2, 4, 8]          & [3, 6, 12, 24]        & [3, 6, 12, 24]        & [4, 8 ,16, 32]              \\ \midrule
  Block Type      & \multicolumn{4}{c}{[DSSA, DSSA\&GSA, DSSA\&GSA, DSSA]}    \\
  \bottomrule
  \end{tabular}}
  \caption{Configurations of SepViT variants.}
  \label{tab:configs}
\end{table}

\subsection{Comparison with Lite Models.}
To further explore the potential of SepViT, we scale down SepViT to a lite model size (SepViT-Lite). As we can observe in Table \ref{tab:Ablation3}, SepViT-Lite obtains an excellent top-1 accuracy of 72.3\%, outperforming its counterparts with similar model sizes.
\begin{table}[h]
  \centering
  \resizebox{0.48\textwidth}{!}{
  \begin{tabular}{c|cc|c}
  \toprule
  Method                & Param(M)   & FLOPs(G)    & Top-1 Acc(\(\% \))       \\  \midrule
  MobileNet-V2 \cite{MobileNet_v2}      & 3.4        & 0.3         & 71.8            \\
  ResNet18 \cite{ResNet}         & 11.1       & 1.8         & 69.8            \\
  PVTv2-B0 \cite{PVT_v2}        & 3.4        & 0.6         & 70.5            \\  
  {\bf SepViT-Lite}                   & 3.7        & 0.6         & {\bf72.3}            \\  
  \bottomrule
  \end{tabular}}
  \caption{Comparison of lite models on classification.}
  \label{tab:Ablation3}
\end{table}

\subsection{Implementation Details}
\subsubsection{ImageNet-1K Classification.}
Image classification experiment is conducted on ImageNet-1K \cite{ImageNet-1K}, which contains about 1.28M training images and 50K validation images from 1K categories.
For a fair comparison, we follow the training settings of the recent vision Transformer \cite{Twins}.
Concretely, all of the SepViT variants are trained for 300 epochs on 8 V100 GPUs with a total batch size of 1024. The resolution of the input image is resized to 224 $\times$ 224.
We adopt the AdamW \cite{AdamW} as the optimizer with weight decay 0.1 for SepViT-B and 0.05 for SepViT-S/T.
The learning rate is gradually decayed based on the cosine strategy with the initialization of 0.001.
We use a linear warm-up strategy with 20 epochs for SepViT-B and 5 epochs for SepViT-S/T.
Besides, we have also employed the increasing stochastic depth augmentation \cite{Stochasticdepth} with the maximum drop-path rate of 0.2, 0.3, 0.5 for our SepViT-T/S/B models.

\subsubsection{Semantic Segmentation.}
Segmentation experiment is carried out on ADE20K \cite{ADE20K}, which contains about 20K training images and 2K validation images from 150 categories.
To make fair comparisons, we also follow the training conventions of the previous vision Transformers \cite{Twins} on the Semantic FPN \cite{Semantic_FPN} and UperNet \cite{UperNet} frameworks. All of our models are pre-trained on the ImageNet-1k and then finetuned on ADE20K with the input size of 512$\times$512. 
For the Semantic FPN framework, we adopt the AdamW optimizer with both the learning rate and weight decay being 0.0001. Then we train the whole networks for 80K iterations with the total batch size of 16 based on the stochastic depth of 0.2, 0.3, and 0.4 for SepViT-T/S/B.
For the training and testing on the UperNet framework, we train the models for 160K iterations with the stochastic depth of 0.3, 0.3, and 0.5. AdamW optimizer is used as well but with the learning rate $6\times10^{-5}$, total batch size 16 and weight decay 0.01 for SepViT-T/S and 0.03 for SepViT-B. Then we test the mIoU based on both single-scale and multi-scale (MS) where the scale goes from 0.5 to 1.75 with an interval of 0.25.

\subsubsection{Object Detection and Instance Segmentation.}
Objection detection and instance segmentation is implemented \cite{COCO} based on the RetinaNet \cite{RetinaNet} and Mask R-CNN \cite{Mask_RCNN} frameworks with COCO \cite{COCO}.
Specifically, all of our models are pre-trained on ImageNet-1K and then finetuned following the settings of the previous works \cite{Twins}.
As for the 12 epochs (1$\times$) experiment, both the RetinaNet-based and the Mask R-CNN-based models use the AdamW optimizer with the weight decay 0.001 for SepViT-T and 0.0001 for SepViT-S.
And they are trained with the total batch size of 16 based on the stochastic depth of 0.2 and 0.3 for SepViT-T/S. 
During the training, there are 500 iterations for warm-up and the learning rate will decline by 10$\times$ at epochs 8 and 11.
For the 36 epochs (3$\times$) experiment with multi-scale (MS) training, models are trained with the resized images such that the shorter side ranges from 480 to 800 and the longer side is at most 1333. 
Moreover, most of all the other settings are the same as the 1$\times$ except that the stochastic depth is 0.3 for SepViT-T, the weight decay becomes 0.05 and 0.1 for SepViT-T/S, and the decay epochs are 27 and 33.

\end{document}